\documentclass[letterpaper, 10 pt, conference]{ieeeconf}  
\IEEEoverridecommandlockouts                              
\title{\LARGE \bf
Vision-Driven 2D Supervised Fine-Tuning Framework for Bird’s Eye View Perception
}

\author{Lei He, Qiaoyi Wang, Honglin Sun, Qing Xu, Bolin Gao, Shengbo Eben Li, Jianqiang Wang, Keqiang Li${^*}$ 
\thanks{*This work  is supported by National Natural Science Foundation of China, Science Fund for Creative Research Groups (Grant No.52221005)}
\thanks{Lei He, Qiaoyi Wang, Honglin Sun, Qing Xu, Bolin Gao, Shengbo Eben Li, Jianqiang Wang, Keqiang Li are with the School of Vehicle and Mobility, Tsinghua University, Beijing, China, and also with the State Key Laboratory of Intelligent Green Vehicle and Mobility, Tsinghua University, Beijing, China (qingxu@tsinghua.edu.cn; lishbo@tsinghua.edu.cn; helei2023@tsinghua.edu.cn; likq@tsinghua.edu.cn )\textit{(Corresponding author: Keqiang Li.)}}%
}

\usepackage{amsmath}  
\usepackage{amssymb}  
\usepackage{graphicx}
\usepackage{comment}
\usepackage{graphicx}
\usepackage{float} 
\usepackage{placeins} 
\usepackage{multirow}  
\usepackage{booktabs}  
\usepackage{pifont}

\renewcommand{\tabcolsep}{6pt}     
\renewcommand{\baselinestretch}{1} 
\renewcommand{\normalsize}{\small} 

\begin{document}

\maketitle
\thispagestyle{empty}
\pagestyle{empty}

\begin{abstract}

Visual bird's eye view (BEV) perception, due to its excellent perceptual capabilities, is progressively replacing costly LiDAR-based perception systems, especially in the realm of urban intelligent driving. However, this type of perception still relies on LiDAR data to construct ground truth databases, a process that is both cumbersome and time-consuming. Moreover, most massproduced autonomous driving systems are only equipped with surround camera sensors and lack LiDAR data for precise annotation. To tackle this challenge, we propose a fine-tuning method for BEV perception network based on visual 2D semantic perception, aimed at enhancing the model’s generalization capabilities in new scene data. Considering the maturity and development of 2D perception technologies, our method significantly reduces the dependency on high-cost BEV ground truths and shows promising industrial application prospects. Extensive experiments and comparative analyses conducted on the nuScenes and Waymo public datasets demonstrate the effectiveness of our proposed method.

\end{abstract}

\section{INTRODUCTION}

With the rapid advancement of artificial intelligence technology, autonomous driving~\cite{1} is swiftly transitioning from the research and experimental stage to large-scale commercial production. The perception module, as one of the core components of autonomous driving systems, is required not only to accurately and promptly perceive the environment around the vehicle but also to respond quickly to dynamic changes to ensure driving safety and stability. The scope of perception technology has continuously expanded, evolving from an initial focus on 2D visual perception to 3D perception, and further developing into the widely adopted BEV perception\cite{2,3}. BEV-based perception frameworks allow the mapping of surrounding environmental information onto a unified 2D plane, enabling a more intuitive and comprehensive understanding of the environment. Combined with the powerful capabilities of Transformers\cite{vaswani2017attention}, the BEV+Transformer\cite{sun2024oe,15,6} perception framework has become the mainstream in the field of autonomous driving, which is able to comprehend both static and dynamic information around the vehicle in real-time, providing a comprehensive 360-degree perception capability for autonomous driving systems. This framework not only enhances the accuracy and efficiency of perception but has also gradually become the foundational technology for end-to-end autonomous driving systems.

However, the development of a perception system based on the BEV+Transformer framework presents significant challenges, chief among them being the reliance on high-precision 4D ground truth data for generating labels that enable effective weight updates in the BEV network\cite{sun2024oe,15,6}. Presently, the automotive industry largely depends on LiDAR data to construct these ground truth datasets. By integrating technologies such as LiDAR and visual SLAM, the static and dynamic elements within the driving environment are geometrically modeled and temporally correlated, with high-precision BEV ground truth data typically generated through manual annotation\cite{7,8,9}. Despite its accuracy, the high cost of LiDAR limits its widespread adoption in mass-produced vehicles, especially in the mid- to low-end segments of the market. Consequently, the majority of production vehicles are equipped only with visual sensors. In challenging scenarios, these vehicles can transmit pure visual data to the cloud, where ground truth data must be constructed. However, the geometric reconstruction process based solely on visual data is highly susceptible to environmental factors such as lighting conditions and surface textures, making accurate ground truth generation exceedingly difficult. This challenge has emerged as a critical technical barrier to the advancement of autonomous driving technology, underscoring the urgent need for innovative solutions to overcome these current limitations.

\begin{figure}[t]  
    \centering
    \vspace{0.3cm}  
   \includegraphics[width=\columnwidth]{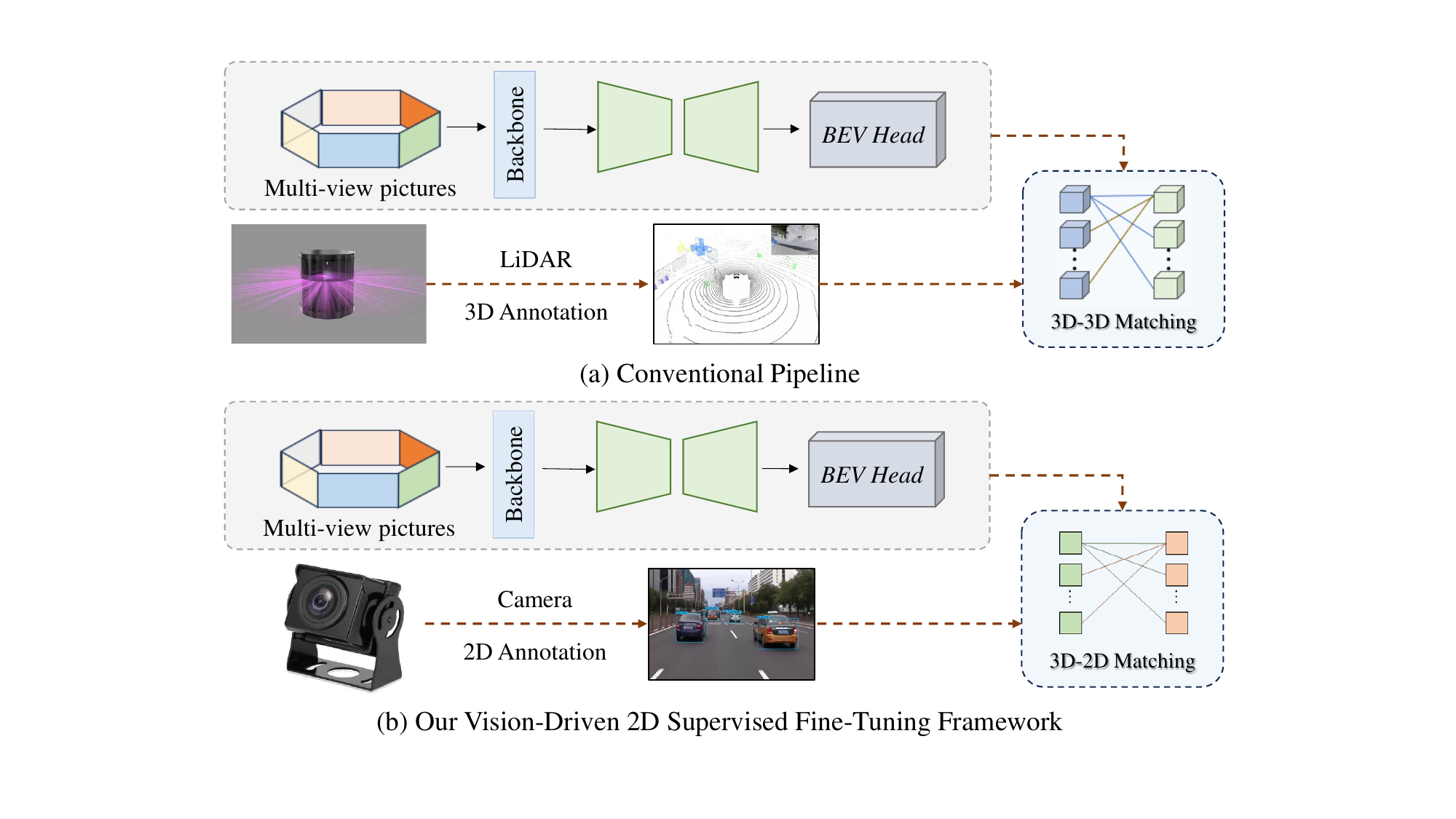}
    \caption{(a) shows the traditional 3D training framework, which relies on LiDAR and 3D annotated data. (b) illustrates our proposed 2D supervised framework, which only requires multi-view images acquisition using purely visual sensors and utilizes 2D annotations to supervise the training of the 3D model, ultimately achieving outstanding performance.}
    \label{fig:1}
\end{figure}


To address the limitations of current ground truth construction methods that rely on expensive LiDAR equipment, we propose a novel approach that fine-tunes the BEV model using 2D image information. Specifically, we begin by manually annotating 2D information in surround-view images or by leveraging a large-scale 2D model for learning. Then, we use the BEV model to infer 3D perception results. These inferred results are subsequently projected onto the surround-view image plane and matched with the existing 2D annotations. This process allows us to construct a loss function, which is then used to further fine-tune the BEV model parameters. This fine-tuning process significantly enhances the generalization capability of the BEV model in new scene data and enables it to generate high-precision 4D ground truth (GT) labels.

Our research makes several key contributions:
\begin{itemize}
\item We propose an innovative method that optimizes the BEV model by utilizing 2D information from surround-view images, thereby improving its adaptability across various complex scenarios. This method provides a low-cost, efficient solution for mass-produced vehicles that lack LiDAR.
\item We design an effective loss function that precisely matches the 3D perception results with 2D annotations, allowing the model to more deeply learn and understand spatial relationships in complex environments. This approach not only enhances the model's accuracy but also improves its ability to handle diverse driving scenarios.
\item We conducted extensive experiments on multiple public datasets, such as nuScenes and Waymo, to validate the effectiveness and superiority of our method. The experimental results demonstrate that the fine-tuned BEV model performs significantly better in diverse scenarios compared to traditional methods, highlighting its great potential for practical autonomous driving applications.
\end{itemize}

\begin{figure*}[t] 
    \centering  
    \includegraphics[width=\textwidth]{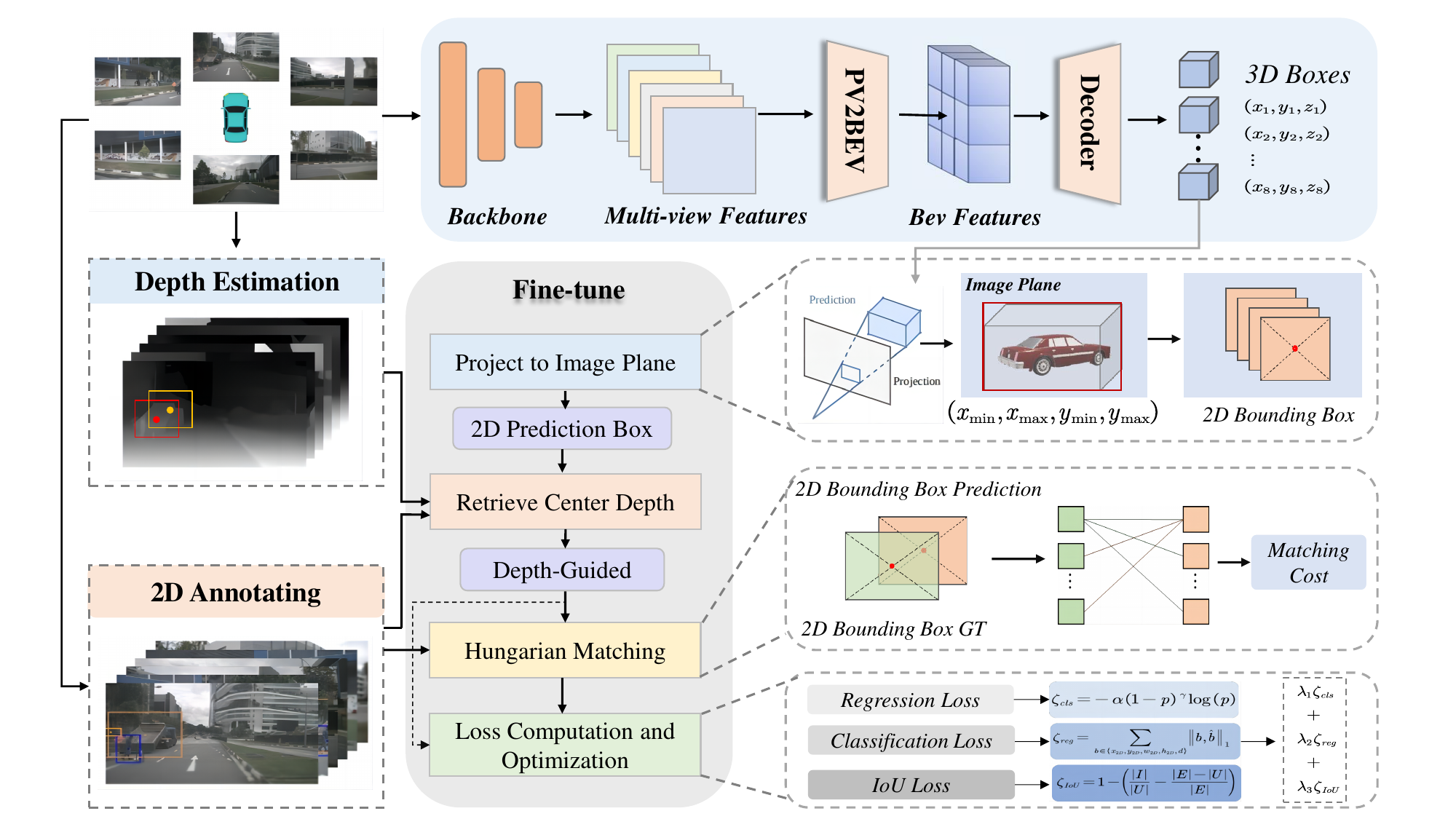}  
    \caption{\textbf{The pipeline of our proposed 2D-supervised fine-tuning model.} The pipeline of our proposed 2D-supervised fine-tuning model is as follows: The 3D perception outcomes are derived through inference by the BEV model, which are then projected onto the plane of the surround-view images for alignment with manually annotated 2D ground truth labels. This alignment is utilized to construct a loss function that facilitates the fine-tuning of the BEV model parameters. Furthermore, depth information is generated offline to assist in the supervision of the matching process, thereby enhancing the accuracy of 3D-2D matching.}
    \label{fig:1}
\end{figure*}

\section{Related Work}

\subsection{Vision-based BEV 3D Object Detection}
Vision-based object detection utilizes cameras as a more cost-effective and straightforward means to help vehicles perceive their surrounding environment. Early 3D object detection methods were typically conducted in perspective view, such as FCOS3D \cite{10}, which employs a single-stage framework to directly predict the 3D attributes of objects. Another monocular 3D detection approach, DD3D \cite{11}, was inspired by pseudo-LiDAR and enhances performance by leveraging pre-trained depth estimation networks \cite{12,13,14}. In recent years, BEV perception has gained widespread attention due to its ability to provide a unified and information-rich 3D scene representation. LSS \cite{15} was one of the early explorations in visual BEV perception, achieving the transformation from perspective view to BEV through depth estimation. Subsequent works, such as BEVDet \cite{16} and BEVDepth \cite{17}, further improved BEV perception by refining depth estimation accuracy. Following this, DETR3D \cite{18}, built on the Transformer framework, facilitated the transformation from perspective view to BEV, drawing on the concepts from DETR \cite{19} to predict bounding boxes. PETR \cite{20}, also based on Transformers, optimized the reference point generation of DETR3D, while BEVFormer enhanced perception by introducing deformable attention \cite{21} and multi-frame fusion techniques. The training and optimization of these methods rely on ground truth generated by LiDAR, yet many production vehicles in the real world primarily depend on vision sensors for environmental perception.

\subsection{3D Object Detection Without Dependence on 3D Supervision}
Due to the high cost associated with precise 3D annotation data, weak supervision, which leverages weakly labeled data such as incomplete or inaccurate annotations \cite{22}, can significantly reduce the training cost of models and accelerate the model development process. The key challenge lies in effectively extracting information from the existing data. Early works like WS3D \cite{23} utilized 2D labels and point cloud data with limited accurate 3D annotations to automatically generate 3D labels for training, thereby effectively reducing the cost of object detection and 3D data annotation. Similarly, FGR \cite{24} achieved 3D vehicle detection through geometric reasoning without relying on 3D labels. WeakM3D \cite{25} also used point cloud data, aligning predicted 2D bounding boxes with the point cloud to obtain Regions of Interest (RoIs), which then assisted the 3D detector in predicting 3D bounding boxes. However, all these methods rely on inputs from multiple modalities, and the requirement for point cloud data implies additional hardware cost.

Furthermore, TDS \cite{26} introduced temporal 2D supervision, using additional 3 seconds of 2D annotations before and after each frame for supervision. Another approach involved generating 3D pseudo-labels through point cloud reconstruction to supervise 3D detectors \cite{27}. These methods integrate 2D supervision into 3D model training, but supervising complex 3D information still requires a small amount of 3D annotations, making it difficult to avoid the need for 3D data to achieve performance comparable to traditional 3D detection models. WSM3D \cite{28} employed inter-frame viewpoint differences for intra-frame 2D supervision, but this method used only monocular 2D supervision for 3D perception, leading to ambiguity issues due to the lack of depth information. Our approach introduces surround-view multi-camera 2D supervision for 3D perception and enhances 3D-2D matching accuracy through offline-generated depth information.

\section{METHOD}

\subsection{Overall Framework} We adopt the advanced 3D object detection framework, BevFormer, and implement an innovative approach to efficiently supervise its training process. Specifically, we utilize manually annotated 2D ground truth bounding boxes and match them with the 2D projections of the 3D bounding boxes output by BevFormer. This matching process is conducted in the image coordinate system and is further supervised by depth information generated offline using DPT\cite{29}. The detailed implementation of this method is illustrated in Figure 2.

\subsection{Generation of 2D Predicted Bounding Boxes}BEVFormer takes multi-view surround camera images as input and interacts within space and time using predefined grid-like BEV queries. Spatial information is aggregated through the Spatial Cross-Attention mechanism, where each BEV query extracts spatial features from the ROI regions in the camera views. Temporal information is integrated through the Temporal Self-Attention mechanism, which iteratively fuses historical BEV information. By leveraging both spatial and temporal information, the model ultimately outputs 3D detection results from the BEV perspective, including predictions of object positions, categories, velocities, and orientations.

The 3D annotations in the nuScenes dataset used by BevFormer are conducted in the LiDAR coordinate system, hence the 3D detection boxes predicted by BevFormer are defined within LiDAR coordinate system. However, the matching process between the predicted boxes and ground truth boxes is performed in the image coordinate system, necessitating the projection of the 3D detection boxes into the image coordinate system. First, the corner points of the 3D bounding boxes $P_L$ output by the 3D Head are transformed into the camera coordinate system using the external parameter matrix. Then, these points are projected onto the image plane using the camera's internal parameter matrix, as shown in Equation 1.

$$
P_I = K \cdot T_{L}^{C} \cdot P_L \eqno{(1)}
$$

Here, $P_L$ represents the coordinates of points in the LiDAR coordinate system,  $T_{L}^{C}$ is the external parameter matrix that transforms coordinates from the LiDAR coordinate system to the camera coordinate system, and $K$ is the internal parameter matrix of the camera. $P_I$ represents the coordinates of points in the image coordinate system.

Through the aforementioned process, we derive the coordinates of the eight corner points of the 3D detection box in the image coordinate system. We then determine the bounding rectangle of the 3D detection box by selecting the maximum and minimum values of these corner points in two different dimensions to form the four corner points of the 2D bounding box, namely \( \left( x_{\min}, x_{\max}, y_{\min}, y_{\max} \right) \). Consequently, we can calculate the width and height of the 2D predicted box as \( w_{2D} = x_{\max} - x_{\min}, h_{2D} = y_{\max} - y_{\min} \). The coordinates of the center point are given by \( x_{2D} = \frac{x_{\min} + x_{\max}}{2} \), \( y_{2D} = \frac{y_{\min} + y_{\max}}{2} \), resulting in the final 2D bounding box \( \left\{ x_{2D}, y_{2D}, w_{2D}, h_{2D} \right\} \).




\subsection{The matching process of 3D training supervised by 2D information}

We have obtained the 2D prediction bounding boxes and, by integrating them with the pre-annotated 2D ground truth bounding boxes in the image coordinate system. The following outlines the specific fine-tuning process.

\subsubsection{\textbf{Incorporating Depth Information}} 
In the projection from 3D to 2D, a 3D bounding box can correspond to a specific 2D bounding box. However, due to the absence of depth information, a given 2D bounding box does not uniquely determine a specific 3D bounding box. In other words, a small object close to the camera and a large object far from the camera may have the same projected 2D bounding box in the image. This can lead to potential matching errors, thereby reducing the model's performance. To mitigate the ambiguity in the 2D-to-3D correspondence, we introduce depth information to assist in the matching of detection boxes.

The depth information for multi-view images is generated offline using the DPT model. DPT utilizes a pre-trained Vision Transformer (ViT) as the backbone to extract multi-level image features, which are then fused through fusion blocks. Finally, the depth map containing depth information is generated through an output convolutional layer.

\subsubsection{\textbf{Hungarian Matching and Loss}} 
For the matching process between the 2D predicted boxes and the ground truth boxes, we take into classification loss, regression loss, and IoU loss. The classification loss measures the discrepancy between the predicted box's class and the ground truth box's class label. The regression loss quantifies the difference between the coordinates of the predicted box and those of the ground truth box. The Intersection over Union (IoU) loss is used to assess the degree of overlap between the two bounding boxes, aiming to enhance the accuracy of the bounding box predictions. By incorporating both 2D bounding box and depth information, we ultimately define the following matching and loss criteria, as shown in Equation 2.

$$
L = \lambda_1 L_{cls} + \lambda_2 L_{reg} + \lambda_3 L_{IoU}\eqno{(2)}
$$

Here, $\lambda_1$, $\lambda_2$, $\lambda_3$ represents the weights for classification loss ($L_{cls}$), regression loss ($L_{reg}$), and IoU loss ($L_{IoU}$). We use Focal Loss\cite{30}, L1 Loss, and GIoU Loss\cite{31}, respectively. The Hungarian algorithm is employed to perform a bipartite matching between the n predicted values and m ground truth values. Similarly, considering the cost of classification, regression, and IoU, the bipartite matching between the predicted values and ground truth pairs $(\pi)$ can be described as shown in Equation 3 and Equation 4:

$$
\pi^* = \mathop{\arg\min}_{\pi} \sum_{i=1}^{n} \sum_{j=1}^{m} c_{ij}\pi_{ij} \eqno{(3)}
$$

$$
C_{ij} = \beta_1 C_{cls} + \beta_2 C_{reg} + \beta_3 C_{Iou} \eqno{(4)}
$$

In this context, $\pi _{ij}$ represents the matching result between the i predicted boxes and the j ground truth boxes obtained by minimizing the matching cost. $\beta_1, \beta_2, \beta_3$ denotes the weights for the classification, regression, and IoU cost respectively. $C_{ij}$ represents the matching cost between the i predicted boxes and the j ground truth boxes, which is the weighted sum of the classification cost $C_{cls}$, the regression cost $C_{reg}$, and the IoU cost $C_{IoU}$.

Due to the significantly higher number of negative samples compared to positive samples in the 2D bounding box matching process, we adopt Focal Loss to address the class imbalance issue by defining the classification loss, as shown in Equation 5.

$$
L_{cls}=-\alpha \left( 1-p \right) ^{\gamma}\log \left( p \right) \eqno{(5)}
$$

In this context, $-\log \left( p \right)$ is the conventional cross-entropy loss function used to calculate the predicted probability, $p$ represents the predicted probability, $\gamma$ is the modulation factor for hard-to-classify samples, which enhances the loss contribution of these challenging samples, and   $\alpha$ is the balancing factor.

The regression loss $L_{reg}$ is defined as the sum of the absolute differences between the predicted and ground truth values for the center point coordinates ($x_{2D},y_{2D}$), width $w_{2D}$, height $h_{2D}$, and depth $d$, as shown in Equation 6.

$$
L_{reg}=\sum_{b\in \left\{ x_{2D},y_{2D},w_{2D},h_{2D},d \right\}}{\left\| b,\hat{b} \right\| _1} \eqno{(6)}
$$

Here, $x_{2D},y_{2D},w_{2D},h_{2D}$ is normalized along the image height direction to ensure consistency in magnitude. It is important to note that the introduction of depth information supervision in the regression loss function aims to enhance the matching accuracy between the predicted and ground truth boxes.

Given that GIoU loss can provide meaningful gradient information, especially in cases where the predicted and ground truth boxes have little to no overlap, it offers better convergence compared to IoU loss. Therefore, we use GIoU loss to represent the IoU loss, as shown in Equation 7.

$$
L_{IoU}=1-\left( \frac{\left| I \right|}{\left| U \right|}-\frac{\left| E \right|-\left| U \right|}{\left| E \right|} \right) \eqno{(7)} 
$$

Here, $I$, $U$, and $E$ represent the intersection, union, and the smallest convex hull region, respectively.

\begin{table*}[t]  
    \centering
    \caption{2D SUPERVISED FINE-TUNING RESULTS ON NUSCENES VALIDATION SET. }
    \label{tab:example}
    \vspace{-0.8 em}  
    \begin{center}
    \begin{tabular}{|c||c||c||c||c||c||c||c|}
        \hline
       \textbf{Method} & \textbf{mAP↑} & \textbf{NDS↑} & \textbf{mATE↓} & \textbf{mASE↓} & \textbf{mAOE↓} & \textbf{mAVE↓} & \textbf{mAAE↓}  \\ \hline
        \textbf{Pre-trained} & 0.2524 & 0.3540 & 0.8976 & 0.2931 & 0.6501 & 0.6557 & 0.2160 \\ \hline
        \textbf{Fine-tuned} & \textbf{0.2775} & \textbf{0.3733} & \textbf{0.8926} & \textbf{0.2908} & \textbf{0.6364} & \textbf{0.6017} & 0.2333 \\ \hline
        
    \end{tabular}
    \end{center}
\end{table*}

\section{Experiments}

\subsection{Dataset and Metrics}

We conducted experiments on two widely recognized public datasets, nuScenes\cite{32} and Waymo\cite{33}, both of which are commonly used benchmarks for 3D object detection tasks. The nuScenes full dataset contains 1,000 scenes, with the training set comprising 700 scenes, the validation set 150 scenes, and the test set 150 scenes. Each scene is annotated with precise 3D bounding boxes at a frequency of 2Hz. The nuScenes training set includes 28,130 samples, and the validation set contains 6,019 samples, with each sample consisting of images captured by six surround-view cameras, covering a 360° field of view. The Waymo Open Dataset comprises 1,150 scenes, with tasks divided into 2D and 3D object detection and tracking. The training set contains 798 scenes, the validation set 202 scenes, and the test set 150 scenes. For the nuScenes dataset, we selected the 10 most common object categories in 3D object detection tasks (car, truck, construction vehicle, bus, trailer, barrier, motorcycle, bicycle, pedestrian, and traffic cone). Similarly, for the Waymo dataset, we chose to detect three object categories (vehicle, pedestrian, and cyclist). Additionally, the 3D detector we employed is BEVFormer-tiny.

We used the nuScenes metrics $mAP$ and $NDS$ to evaluate the model's performance on both datasets. $NDS$ is a combination of $mAP$ and various TP metrics, with the combined calculation method shown in Equation 8.

$$
\label{eq:nds}
    \text{NDS}=\frac{1}{10}[5\cdot\text{mAP}+\sum_{\text{mTP} \in \mathbb{TP}}(1-\min(1,\text{mTP}))] \eqno{(8)} 
$$

Among these,$ mTP$ is a composite metric that includes the following five indicators: mean Average Translation Error (mATE), mean Average Scale Error (mASE), mean Average Orientation Error (mAOE), mean Average Velocity Error (mAVE), and mean Average Attribute Error (mAAE). mAP represents the mean Average Precision, with a weight of 5, and is used to measure the overall detection accuracy of the model across all categories. The calculation method for mAP in nuScenes is shown in Equation 9.

$$
mAP=\frac{1}{\left| \mathbb{C} \right|\left| \mathbb{D} \right|}\sum_{c\in \mathbb{C}}{\sum_{d\in \mathbb{D}}{AP_{c,d}}} \eqno{(9)} 
$$

Here, $\mathbb{C}$ represents the set of target categories, and $\mathbb{D}$ denotes the set of matching thresholds, $\mathbb{D}= \{0.5, 1, 2, 4\}$ meters, which is applied across all categories. AP represents the Average Precision for category $c$ and matching threshold $d$.

$\sum_{\text{mTP} \in \mathbb{TP}}(1-\min(1,\text{mTP}))$ is the sum of penalty terms for mTP. For each TP metric, if mTP is less than 1, the penalty term is calculated as $1-\text{mTP}$. If mTP is greater than or equal to 1, the penalty term is 0. Since mAVE (mean Average Velocity Error), mAOE (mean Average Orientation Error), and mATE (mean Average Translation Error) may exceed 1, this approach constrains each metric to a range between 0 and 1.
\subsection{Experimental Settings}
\subsubsection{\textbf{Pre-train}} 
First, we performed pre-training of the 3D detection model on the training sets of both the nuScenes and Waymo datasets. For the model pre-training on nuScenes, we utilized the officially released BEVFormer-tiny weights. The pre-training on the Waymo Open Dataset followed the same training strategy as used for BEVFormer on nuScenes. We employed the AdamW optimizer and set the maximum learning rate to 2e-4 with a cosine annealing learning rate schedule, training for 24 epochs.

\subsubsection{\textbf{Implementation Details}} 
Before fine-tuning, we used a pre-trained DPT-hybrid model to generate depth maps offline, facilitating the subsequent incorporation of depth information from predicted and ground truth bounding boxes during training.

The fine-tuning process was conducted on the validation sets of both the nuScenes and Waymo datasets. We used validation sets containing only 2D annotations for supervision, simulating the training process with newly collected data. During fine-tuning on the Waymo dataset, when calculating NDS, the values of mAVE and mAAE were set to 1. The maximum learning rate for the cosine annealing schedule was set to 2e-6, and the training was conducted for 6 epochs. The learning rates of the backbone and neck are set to 0.1 and 0.5 times., respectively. The coefficients for the classification, regression, and IoU loss were set to $\lambda_1=2, \lambda_2=0.75$, and $\lambda_3=0.25$.  The coefficients for the classification, regression, and IoU cost were set to $\beta_1=2, \beta_2=0.75$, and $\beta_3=0.25$. The parameters for Focal Loss were set to $\alpha=0.25$ and $\gamma=2$.

During fine-tuning, to prevent the model from forgetting previously learned 3D knowledge or overfitting on the validation set, we employed a joint training strategy. Specifically, we combined the training set of the nuScenes dataset, which contains 3D labels, with the validation set, which only contains 2D labels. In this setup, the model is simultaneously supervised using 3D labels on the training set and 2D labels on the validation set. It is important to note that the 3D labels mentioned here are predefined. Even when applying our model to datasets that only contain surround-view camera images, we only need to manually annotate 2D labels and can still leverage the 3D labels from the nuScenes dataset for joint training. This approach effectively enhances the generalization capability of the training process while avoiding the loss of knowledge.

\begin{table}[t]  
    \centering
    \caption{2D SUPERVISED FINE-TUNING RESULTS ON WAYMO VALIDATION SET.}
    \label{tab:example}
    \resizebox{\columnwidth}{!}{  
    \begin{tabular}{|c||c||c||c||c||c|}
        \hline
        \textbf{Method} & \textbf{mAP↑} & \textbf{NDS↑} & \textbf{mATE↓} & \textbf{mASE↓} & \textbf{mAOE↓} \\ \hline
        \textbf{Pre-trained} & 0.2979 & 0.2615 & 0.8169 & 0.4753 & 0.5821 \\ 
        \hline
        \textbf{Fine-tuned} & \textbf{0.3100} & \textbf{0.2693} & \textbf{0.8061} & \textbf{0.4752} & \textbf{0.5761} \\ \hline
    \end{tabular}
    }
\end{table}

\subsection{Benchmark Results}
Table I presents the experimental results of fine-tuning on the nuScenes validation set. Starting from the pre-trained model, we fine-tuned it using the 2D labels from the validation set. Evidently, our method improved the mAP from 0.2524 to 0.2775, representing an increase of 2.51 percentage points; the NDS metric increased from 0.3540 to 0.3722, showing an improvement of 1.93 percentage points; the mATE metric decreased from 0.8976 to 0.8926, improving by 0.5 percentage points; the mASE metric decreased from 0.2931 to 0.2908, improving by 0.23 percentage points; the mAOE metric decreased from 0.6501 to 0.6364, improving by 1.37 percentage points; and the mAVE metric decreased from 0.6557 to 0.6017, reflecting an improvement of 5.4 percentage points. The mAAE (mean Average Attribute Error) metric slightly increased, indicating a minor rise in the average error when predicting object attributes, but this had a negligible impact on the overall model performance. Overall, the model exhibited varying degrees of improvement across several key metrics, demonstrating that fine-tuning with the 2D labels from the nuScenes validation set effectively enhanced the model's performance.

Table II presents the results of 2D supervised fine-tuning on the Waymo Open Dataset[30] validation set. Due to the lack of certain object velocity and attribute data in the Waymo dataset, it was not possible to accurately calculate mAVE (mean Average Velocity Error) and mAAE (mean Average Attribute Error). To facilitate the calculation of NDS, we set the values of mAVE and mAAE to 1. Therefore, Table II evaluates only the three TP metrics: mATE, mASE, and mAOE. Obviously, the mAP increased from 0.2979 to 0.3100, representing an improvement of 1.21 percentage points; the NDS increased from 0.2615 to 0.2693, an improvement of 0.78 percentage points; the mATE metric decreased from 0.8169 to 0.8061, improving by 1.08 percentage points; the mASE metric decreased from 0.4753 to 0.4752, showing a slight improvement of 0.01 percentage points; and the mAOE metric decreased from 0.5821 to 0.5761, improving by 0.6 percentage points. Overall, although it was not possible to accurately assess the mAVE and mAAE metrics, the model demonstrated varying degrees of improvement across the other evaluable metrics. This indicates that the 2D supervised fine-tuning process also enhanced the model's performance on the Waymo Open Dataset validation set, proving the effectiveness of our proposed method.

\begin{table*}[t]  
    \centering
    \caption{ABLATION STUDY ON NUSCENES VALIDATION SET.}
    \label{tab:example}
    \vspace{-0.8 em}  
    \begin{center}
    \begin{tabular}{|c||c||c||c||c||c||c||c||c||c||c|}
         \hline
        {\textbf{Method}} 
        & {\textbf{DPT}} 
        & {\textbf{IOU}} 
        & {\textbf{Joint}} 
        & {\textbf{mAP}} 
        & {\textbf{NDS}} 
        & {\textbf{mATE}} 
        & {\textbf{mASE}} 
        & {\textbf{mAOE}} 
        & {\textbf{mAVE}} 
        & {\textbf{mAAE}} \\ 
        \hline
        \multirow{4}{*}{\textbf{2D supervision}} 
        &  &  &  & 0.2611 & 0.3547 & 0.9175 & 0.2920 & 0.6554 & 0.6573 & 0.2361 \\ \cline{2-11}
        & \ding{51} &  &  & 0.2685 & 0.3601 & 0.9200 & 0.2974 & 0.6373 & 0.6566 & 0.2299 \\ \cline{2-11}
        & \ding{51} & \ding{51} &  & 0.2724 & 0.3619 & 0.9078 & 0.3011 & 0.6474 & 0.6634 & 0.2232 \\ \cline{2-11}
        & \ding{51} & \ding{51} & \ding{51} & 0.2775 & 0.3733 & 0.8926 & 0.2908 & 0.6364 & 0.6017 & 0.2333 \\ \hline
    \end{tabular}
    \end{center}
\end{table*}


\subsection{Ablations and Analyses}
In this section, we incrementally add depth estimation, GIoU loss, and joint training to the baseline model, verifying the effectiveness of each module through ablation experiments.

The results are shown in Table III, where DPT represents the depth information of the predicted and ground truth boxes, IOU denotes the GIoU Loss, and JOINT refers to joint training. Incorporating depth estimation improved the mAP metric from 0.2611 to 0.2685, an increase of 0.74 percentage points, and raised the NDS from 0.3547 to 0.3601, an improvement of 0.54 percentage points. Introducing GIoU Loss further enhanced the mAP from 0.2685 to 0.2724, an increase of 0.39 percentage points, and improved the NDS from 0.3601 to 0.3619, an increase of 0.18 percentage points. Joint training improved the mAP from 0.2724 to 0.2775, an increase of 0.51 percentage points, and raised the NDS from 0.3619 to 0.3733, an improvement of 1.14 percentage points. Even when using only 2D supervision, the mAP showed improvement over the pre-trained model; however, due to the lack of 3D information, most TP metrics representing 3D performance slightly declined. Incorporating depth information and GIoU Loss further increased the mAP, but with minor fluctuations in TP metrics. Joint training, which integrates the 3D annotations from the training set with 2D annotations from the validation set, proved effective in maintaining 3D performance. Without joint training, the lack of supervision for 3D information such as orientation and velocity led to poorer TP metrics. By mixing fine-tuning data with the training set that includes 3D annotations, we mitigated this issue and observed improvements in multiple TP metrics. This validates the effectiveness of joint training in preserving 3D knowledge and enhancing overall model performance.

\begin{figure*}[t]  
    \centering  
    \includegraphics[width=\textwidth]{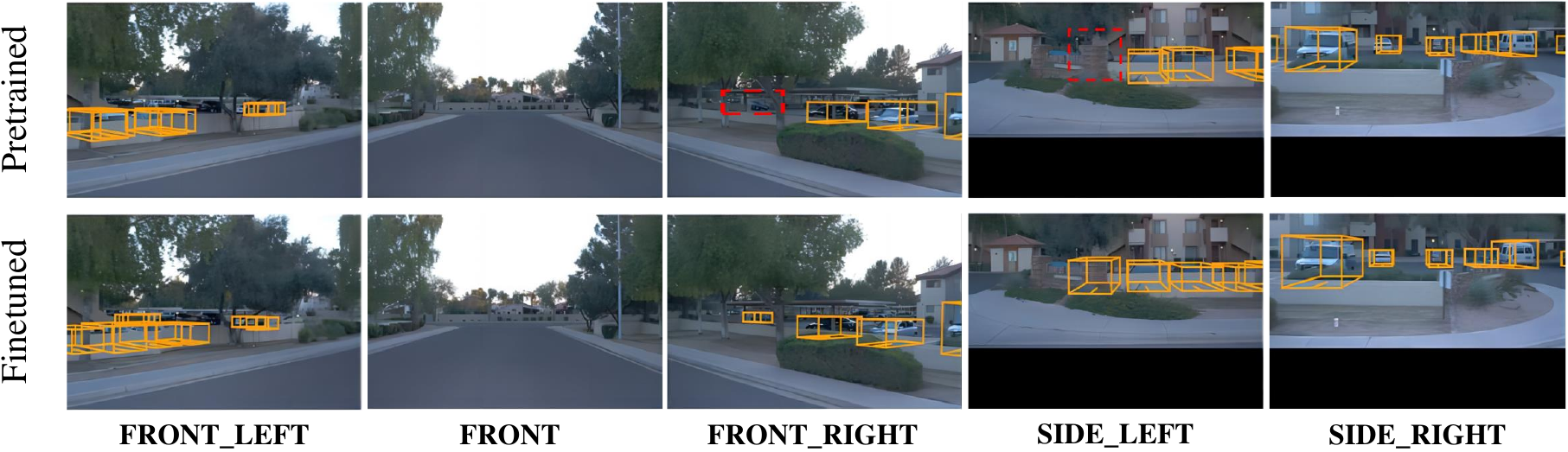}  
    \caption{Visualization of fine-tuning results on the Waymo dataset. We display the 3D predictions from five different viewpoint images. The first row shows the predictions from the pre-trained model, while the second row presents the predictions after fine-tuning.}
    \label{fig:1}
\end{figure*}

\begin{figure*}[t]  
    \centering  
    \includegraphics[width=\textwidth]{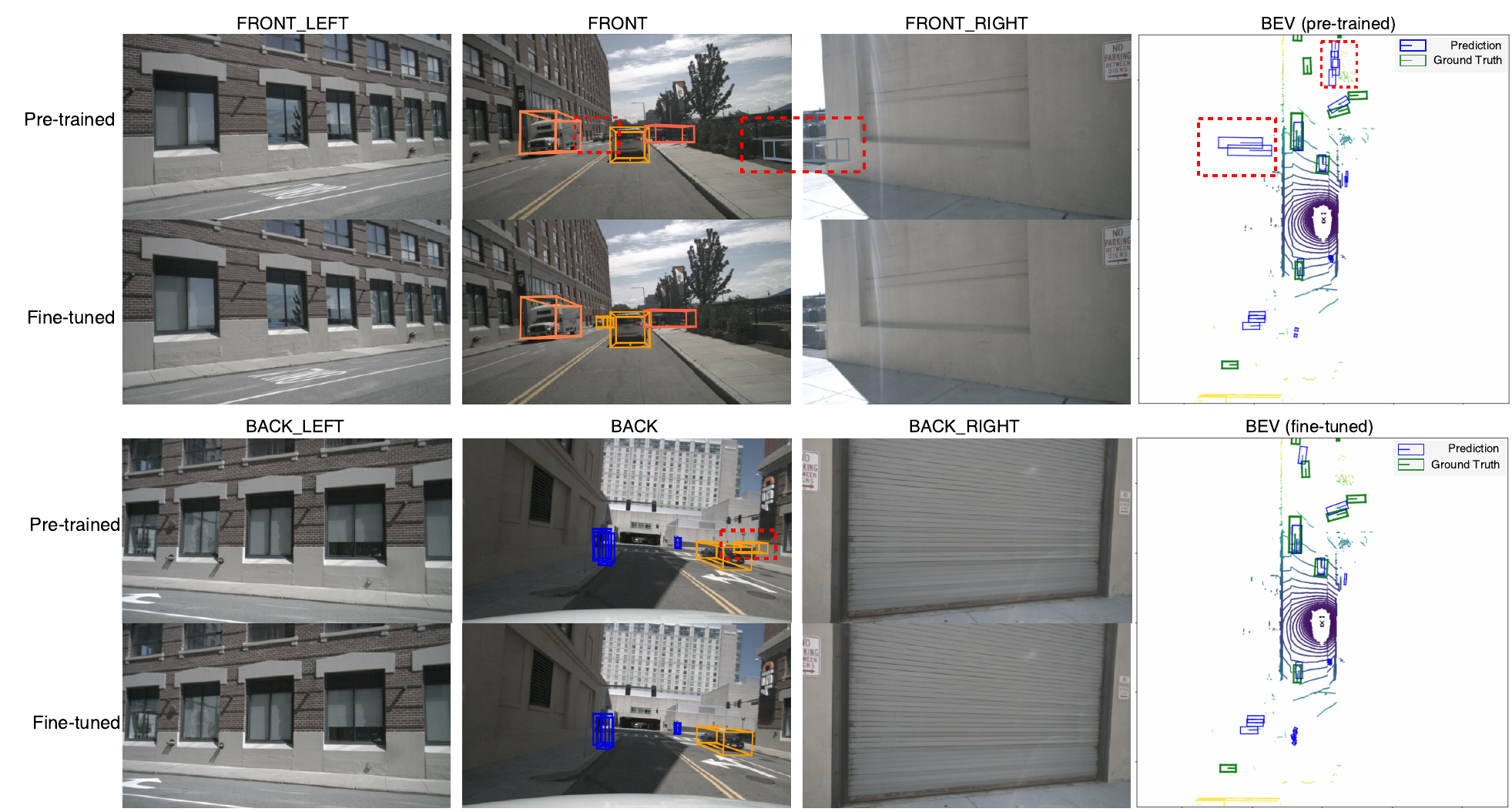}  
    \caption{Visualization of fine-tuning results on the nuScenes dataset. The first three columns display 3D predictions from six different viewpoint images, while the fourth column presents the pre-training and fine-tuning detection results from the BEV perspective for the corresponding scene. The first two rows show predictions from the pre-trained model, and the last two rows present predictions from the fine-tuned model.}
    \label{fig:1}
\end{figure*}

Comparing the fine-tuning results of the baseline with those obtained after adding depth estimation, GIoU loss, and joint training, we observed that the mAP improved from 0.2611 to 0.2775, a gain of 1.64 percentage points, and the NDS increased from 0.3547 to 0.3733, a gain of 1.86 percentage points. These significant improvements in both key metrics, along with the various degrees of enhancement in other TP metrics, underscore the overall efficacy of our approach.

These results indicate that through 2D supervised training, the model can effectively learn from new scenes, leading to improvements in metrics such as mAP and NDS. Furthermore, the introduction of joint training plays a crucial role in maintaining 3D detection performance, further enhancing the effectiveness of 2D supervision. The experimental results above validate the effectiveness and feasibility of our proposed approach for training 3D models using 2D labels, without relying on 3D annotations.

\subsection{Visual Comparisons}
This is the visualization of the fine-tuning results for a specific scenario in the Waymo dataset. Since the Waymo dataset only contains five surround-view camera perspectives, the images shown in the figure are limited to detection results from these five perspectives.

As shown in Figure 3, this is a visualization of the fine-tuning results for a specific scene in the Waymo dataset. Since the Waymo dataset includes only five surround-view camera perspectives, the images shown in the figure are limited to detection results from these five perspectives. Comparing the detection results between the pre-trained model and the fine-tuned model, the areas marked with red boxes highlight significant instances of missed detections due to occlusion, which were ultimately detected after applying the fine-tuning method.

As shown in Figure 4, we conducted a visual analysis of a randomly selected scene from the nuScenes dataset. In the visualization of the pre-trained model's results, we used red dashed boxes to highlight significant instances of false positives and missed detections. For example, in the front view and front-right view of this scene, the pre-trained model produced false positives marked by gray boxes, and there were also instances of missed detections in the front view. However, in the fine-tuned model, these false positives and missed detections were effectively mitigated, demonstrating significant improvements after fine-tuning. From the BEV view, it is also evident that our model effectively avoided false positives and missed detections. These visualizations confirm the effectiveness of our proposed method.
\section{CONCLUSIONS}
In this paper, we proposed a novel approach to fine-tune the BEV model using 2D annotations from surround-view cameras. By projecting the 3D perception outputs of the BEV model onto the image plane and matching them with manually annotated 2D labels and corresponding depth estimates, we constructed a loss function to optimize the BEV model parameters. Unlike traditional BEV models that depend on LiDAR for generating 3D ground truth, our method leverages 2D and depth estimation supervision, significantly enhancing runtime efficiency. Experimental results on the nuScenes and Waymo datasets demonstrate the superior performance of our approach, marking the first successful validation of constructing BEV perception ground truth solely using visual sensors without LiDAR. Looking forward, we plan to extend this surround-view camera-supervised BEV method to additional tasks such as local map construction and occupancy network perception, further broadening its applications.





\end{document}